**Generative deep learning for foundational video translation in ultrasound**

**Authors:**

Nikolina Tomic MS[1], Roshni Bhatnagar MD[1], Sarthak Jain[1], Connor Lau[1], Tien-Yu Liu MS[1], Laura Gambini PhD[1], Rima Arnaout MD[2†]

[1]Department of Medicine, Division of Cardiology

Bakar Computational Health Sciences Institute

University of California, San Francisco

San Francisco, CA 94143

[2]Department of Medicine, Division of Cardiology

Bakar Computational Health Sciences Institute

UCSF-UC Berkeley Joint Program in Computational Precision Health

Department of Radiology, Center for Intelligent Imaging

University of California, San Francisco

† Corresponding Author

**Word Count:** 4129

**Abstract**

Deep learning (DL) has the potential to revolutionize image acquisition and interpretation across medicine, however, attention to data imbalance and missingness is required. Ultrasound data presents a particular challenge because in addition to different views and structures, it includes several sub-modalities—such as greyscale and color flow doppler (CFD)—that are often imbalanced in clinical studies. Image translation can help balance datasets but is challenging for ultrasound sub-modalities to date. Here, we present a generative method for ultrasound CFD-greyscale video translation, trained on 54,975 videos and tested on 8,368. The method developed leveraged pixel-wise, adversarial, and perceptual loses and utilized two networks: one for reconstructing anatomic structures and one for denoising to achieve realistic ultrasound imaging. Average pairwise SSIM between synthetic videos and ground truth was 0.91±0.04. Synthetic videos performed indistinguishably from real ones in DL classification and segmentation tasks and when evaluated by blinded clinical experts: F1 score was 0.9 for real and 0.89 for synthetic videos; Dice score between real and synthetic segmentation was 0.97. Overall clinician accuracy in distinguishing real vs synthetic videos was 54±6% (42-61%), indicating realistic synthetic videos. Although trained only on heart videos, the model worked well on ultrasound spanning several clinical domains (average SSIM 0.91±0.05), demonstrating foundational abilities. Together, these data expand the utility of retrospectively collected imaging and augment the dataset design toolbox for medical imaging.

## Introduction

Ultrasound is one of the highest-volume medical imaging modalities worldwide and is critical to diagnosis and management across organ systems. Deep learning (DL) has numerous use cases in ultrasound, from cardiac imaging[1–5] to fetal screening[6,7] to abdominal evaluation[8]. DL can expand access to expert image interpretation[6,7], extract additional information from existing images[3,9,10], and improve efficiency[11,12] of interpretation, segmentation, and other tasks[1,13,14].

Growing experience has shown the importance of curating training and test datasets[15–17] for DL models. For example, it is increasingly clear that proper dataset design must consider overall dataset diversity, including both inter- and intra-class variation, above and beyond the simple metrics of size and class balance[18,19]. Similarly, while off-the-shelf data augmentation strategies improve training-time regularization and performance, strategies that leverage domain-specific structure and content knowledge about clinical imaging can provide additional value[20]. Without dataset design and augmentation strategies that recognize and balance and/or mitigate potential inter- and intra-class confounders, DL models may fail to learn clinically relevant features for the task at hand, potentially overfitting on artifacts or confounding features that happen to differ among classes.

Ultrasound imaging exemplifies these challenges, in large part because they comprise several sub-modalities including B-mode, color flow Doppler (CFD) (**Fig. 1A**), m-mode, spectral Doppler, and more[21]. In particular, B-mode and CFD both display two-dimensional images/videos of anatomic structures and are therefore likely to be considered as a single modality. However, the use of CFD varies greatly by sonographer, protocol, views and anatomic structures being imaged, leading to an imbalanced presence of CFD within classes of interest for many important clinical tasks[1,6] (**Fig. 1C-1F**). Thus, the presence of CFD signals in an image can easily be misconstrued as a feature of interest in classification, segmentation, or other tasks. At the same time, avoiding this potentially confounding variable by removing all CFD images from the dataset discards valuable data and worsens class imbalance.

Instead, the ability to translate between CFD and B-mode would be a useful way to balance use of CFD among classes without discarding data. Removing the CFD signal from an image would seem to be a simple task, but in reality, belies simple algorithmic methods due to variable CFD color palettes, variable B-mode color maps (chromamaps), color mixing, variable gain as well as noise profiles in the underlying ultrasound image, and obfuscation of underlying anatomic structures (**Fig. 1B**).

DL-based image (or video) translation and inpainting can potentially overcome these challenges. Image translation has been applied to medical imaging in various cases, such as translating between MRI sub-modalities[22,23], MRI motion correction, PET denoising, and PET-CT translation[24]. Similarly, inpainting missing or degraded image regions has been widely studied for natural images[25–28], but in the medical domain, has largely remained limited to inpainting uniformly-shaped, artificially generated patches on individual frames[29,30]. CFD to B-mode translation must balance style transfer, content preservation, realism, and reconstruction of free-form regions with anatomic structures that are not seen in the CFD image/video. Synthetic images/videos must be robust to rigorous evaluation[31], especially in clinical domains. Ideally, the approach should work effectively on ultrasounds from all organ systems, even if not explicitly trained on them.

Previous deep learning methods have addressed some of the above needs, but for photographic/natural imaging and in piecewise fashion. For example, generative adversarial networks (GANs) are common architectures for image and video translation[32]. GANs consist of a generator and a discriminator, where the generator is trained using both per-pixel and adversarial training losses. When translation also requires inpainting, however, a more complex model and training strategy is needed to accurately reconstruct missing regions and maintain visual and semantic consistency with the surrounding image[25–27,29,30]. Consequently, additional non-adversarial losses[33] and gated convolutions [24] have been utilized to enable semantically consistent inpainting. (As networks based on gated convolutions prioritize structural coherence over fine detail, a secondary network is typically used to recover fine textures[27,30].) To date, however, there has not been a method that addresses all specifications for ultrasound, and especially not in a way that is robust to functional and human perceptual testing by clinicians.

Here, we present a generative method for video translation from CFD to B-mode ultrasound that leverages pixel-wise, adversarial, and perceptual losses to both reconstruct anatomic structures and achieve realistic ultrasound textures. We demonstrate its effectiveness across different anatomic structures: adult cardiac ultrasound, fetal obstetric ultrasound, and zero-shot performance on additional anatomic structures. We report performance using algorithmic metrics and several functional tests with benchmark DL models and clinical experts. The model demonstrates foundational ability to perform video translation on ultrasounds from a range of clinical domains not seen during training.

## Results

We developed a generative deep learning pipeline to translate CFD ultrasound video to B-mode (**Fig. 2**, Methods). We trained and validated on 54,975 videos across fetal and adult ultrasounds with real-world clinical heterogeneity. We tested on 8,370 videos (2,866 adult and 5,504 fetal, **Table S1**) as well as ultrasound from several other clinical domains.

Overall, the structural similarity index measure (SSIM) between ground-truth B-mode frames and their synthetic counterparts was 0.91±0.04 (0.1-0.98, n=8370, **Fig. 3-4**). The Fréchet Inception Distance (FID) was 36.

Synthetic videos were then evaluated using several functional tests. The performance of synthetic videos in benchmark view classification[4,6] and segmentation[1] tasks was statistically similar to that of real ones (**Fig. 5**). Clinical experts asked to distinguish real videos from synthetic ones achieved only 54±6% accuracy overall (range, 42-61%) —no better than random chance.

Finally, to assess foundational ability of our approach, we performed zero-shot evaluation on ultrasound exam types the model was not trained on and achieved an average SSIM of 0.91±0.05 (0.43-0.97, n=260, **Fig. 6**, **Table S2**). Taken together, these results indicate realistic, clinically relevant video-to-video translation among ultrasound sub-modalities.

### *Structural similarity of synthetic video to ground truth is high*

A minority of ultrasound acquisitions have simultaneously acquired B-mode and CFD videos. We leveraged these dual ultrasound acquisitions to train the GAN pipeline in a supervised manner, where the B-mode panel served as the ground truth. To ensure balanced learning across varying levels of CFD coverage, we quantified the proportion of CFD coverage within each video and divided the data into color quartiles (Q1–Q4). The model was guided to attend to CFD regions and trained in a two-step pipeline: a coarse network performing anatomic reconstruction, followed by a refinement network guiding style and texture.

We compared SSIM values between real B-mode and CFD image pairs, and between real B-modes and their synthetic B-mode counterparts. The resulting Wasserstein metrics showed notable SSIM shifts across adult, fetal, and other anatomic datasets (0.07, 0.03, and 0.02; **Table S3**), indicating that the synthetic B-mode outputs differ from the CFD inputs and more closely resemble the B-mode ground truth.

Since greater CFD coverage corresponds to a larger portion of the video frames requiring reconstruction, we analyzed performance across the previously defined quartiles (examples in **Fig. 3-4,6**; see also Methods, Estimating CFD Masks). As expected, higher CFD coverage was associated with slightly lower average SSIM yet higher Wasserstein distance. This was most prominently observed in Q4, which exhibited the largest deviation from input CFD videos.

*Synthetic videos performed indistinguishably in benchmark DL tasks*

In addition to SSIM and FID, we also evaluated the performance of synthetic B-mode vs. real B-mode videos in two benchmark DL tasks—view classification and semantic segmentation—on two distinct datasets: adult and fetal ultrasound.

The F1 score on real B-mode videos (n=2239) in an adult view classifier[4] was 0.9, while the F1 score using synthetic B-mode videos was 0.89. There was no statistical difference among per-class F1 scores between real and synthetic videos (p=0.68). The overall F1 score on real B-mode videos (n=2203) in a fetal view classifier[6] was 0.8, while the F1 score using synthetic B-mode videos was 0.79. As with the adult view classification task, there was no statistical difference among per-class F1-scores between real and synthetic videos (p=0.81). Confusion matrices comparing inference of real vs synthetic videos for each task (**Fig. 5A-B**) had F1 scores of 0.95 and 0.93 for adult and fetal comparisons, respectively.

We then leveraged benchmark DL models for semantic segmentation for adult and fetal ultrasound[1,6] (**Fig. 5C-D**). We compared the segmentation from real frames with segmentations from their synthetic B-mode counterparts. The average Dice score per class (excluding background) was 0.97±0.03, for adult frames (range 0.87-0.99, n=99) and 0.9±0.16 for fetal frames (range 0.1-0.99, n=172).

Taken together, these data show that real and synthetic videos are indistinguishable to benchmark DL models.

*Clinicians found real and synthetic videos indistinguishable*

Beyond SSIM and functional testing in DL models, we evaluated human perception of synthetic videos using clinical experts.

*Human perception on clinical view classification.* Clinical experts in adult (n=6) and fetal (n=8) ultrasound were asked to perform view classification—a common task in clinical practice— while blinded to the fact that some videos were synthetic. A post-hoc analysis of human view classification showed indistinguishable performance on real vs synthetic videos (p=0.26 and p=0.27 for adult and fetal, respectively). **Table 1** details this performance.

*Human perception directly regarding video realism.* Separately, clinicians (n=3 for each adult task, n=4 for each fetal task) were asked to identify if a given video was real or synthetic (i) when given one video at a time, and (ii) when given a pair of videos in which they are told one is real and one is synthetic.

On the first task, accuracy was 54% for adult and 57% for fetal videos (where 50% accuracy represents random chance). On the second task, in which paired examples of real and fake videos were provided, accuracy was 47% for adult, and 55% for fetal videos. Overall, accuracy was 54±6% (range, 42-61%). **Table 1** details this performance by task and dataset.

Independently, three participants were asked to identify whether each synthetic video (n=100) was real or synthetic, and for each sample, a fool rate coefficient was computed representing the proportion of subjects who perceived it as real. We analyzed the correlation between fool rate and SSIM between real B-mode and synthetic video. $R^2$ was 0.013 for adult and 0.168 for fetal data within the observed SSIM range (0.92±0.03 range 0.83-0.96 for adult; 0.93±0.03, range 0.84-0.98 for fetal).

<u>*Balancing Fetal Ultrasound training datasets with synthetic data*</u>

Frames extracted from synthetic videos were also used to improve yield, class balance, and diversity[11] of fetal imaging data used for clinical congenital heart disease (CHD) detection from B-mode imaging[6]. Specifically, 26,687 frames from 456 ultrasound exams (2100 video clips in total) were recovered through our generative translation approach. Previously, we had found that most diversity in fetal imaging data resides at the video clip level[11]; generative translation recovered 7.1% more video clips for clinical model training.

These data included imaging from 12 external sites with varying percentages of CFD imaging available (Fig 1F). By recovering more B-mode imaging through our generative translation pipeline, the representation of the most under-represented clinical site increased by 19.8%. Similarly, the there was modest improvement in view class balance. Finally, regarding disease distribution (normal vs congenital heart disease) generative translation recruited significantly more of the most underrepresented CHD lesions (38% to 44%).

It should be noted that the above impacts of generative translation in improving clinical imaging dataset quality and content[34] are significant underestimates, since, due to the lack of a powerful image translation pipeline previously, we have systematically avoided annotation of CFD imaging to date.

*Zero-shot performance on ultrasound covering other anatomies*

To access the foundational ability of our approach, we performed zero-shot evaluation on 260 additional ultrasound exams (**Table S2**) spanning 11 different anatomies—ultrasound types the model was not trained on. Across this diverse dataset, average SSIM was 0.91±0.05 (0.43-0.97, n=260) and the FID was 78. After visual assessment, we found that the synthetic videos generally align well with real B-mode videos (**Fig. 6**).

Radiologists specializing in ultrasound were asked to classify anatomy (n=6) and complete two tasks (n=3 for each task) to determine whether each video (n=100 for each task) was real or synthetic, as above. A post-hoc analysis of anatomy classification showed indistinguishable performance on real vs synthetic videos (with a trend toward significance, p=0.06).

On the tasks to identify if a given video was real or synthetic, accuracy was 52% when presented single video, and 79% when presented a pair of videos (**Table 1**).

## Discussion

Deep learning has shown its potential to be a powerful tool in revolutionizing clinical ultrasound. Nevertheless, the existence of several, often imbalanced sub-modalities can hinder curation of the diverse and balanced datasets needed to develop robust and generalizable DL models for medicine.

Here, we introduce a video-to-video generative translation method that effectively translates two commonly conflated ultrasound sub-modalities, B-mode and CFD. A strength of our study is demonstrated robust performance on diverse ultrasound datasets, including on videos with a high degree of CFD signal, requiring significant reconstruction of underlying anatomic structures. Our findings further illustrate the practical utility of generative image translation in improving the yield, diversity, and class balance of such datasets for use in clinical DL models.

Another strength is the thorough evaluation of synthetic videos, with several computational and blinded clinician perception evaluations of synthetic data quality, where synthetic medical imaging must be both realistic and clinically accurate. This includes good performance on 11 anatomic structures not included in model training, thus demonstrating the pipeline's foundational capabilities for video translation.

We found that synthetic videos were statistically indistinguishable from real ones, across computational measures, functional measures in deep learning models, and using clinical human perception. From this cross-functional testing, we also explore a critical issue in the wider field of generative artificial intelligence (AI): that of how best to evaluate AI-generated imaging, especially in high-stakes clinical domains. Generative models trained in a supervised fashion offer an ability to evaluate instance-to-instance comparisons; in this case, structural similarity between the ground-truth and synthetic B-mode frames. SSIMs between ground truth and synthetic frames were high as an absolute measure and relative to B-mode-CFD and random pairwise B-mode SSIMs (**Figs. 3-4**, **Table S3**), indicating excellent performance.

Furthermore, synthetic videos were highly realistic in human perceptual testing. Clinicians given a single video could not reliably distinguish whether it was real or synthetic, even when presented with a pair of real and synthetic videos side by side. Additionally, weak to moderate overall correlation between SSIM and human perception was found, but a stronger relationship was observed at lower SSIM values, where visual differences become more pronounced. These findings align with previous literature showing that traditional image quality metrics like SSIM, while widely used in natural imaging tasks, have limits when capturing perceptual and diagnostic quality relevant to clinical interpretation[35,36].

We found that our model demonstrated foundational capabilities for image translation as it operates across multiple views and anatomical structures on which it was not trained, while acknowledging that the definition of a foundation model remains open[13]. The generative model's ability to translate imaging in ultrasound is already more generalized than the clinical experts asked to judge the model's outputs; while the model operates across adult and fetal use cases, clinical experts are specialized in one area or another.

While generative translation was excellent overall across several datasets and performance metrics, there were limitations to our approach. We observed cases where the synthetic videos exhibited noticeable blurriness and occasional artifacts. Thus, perceptual quality remains an area for improvement in our pipeline. This was slightly worse for fetal data compared to adult, reflected in performance metrics. There is a general consensus among experts that fetal imaging is more difficult than adult due to small fetal size and unpredictable motion; furthermore, the use of CFD in the small fetal heart is more likely to encompass the entire heart, obfuscating underlying structures more thoroughly than in adult. Indeed, even on real B-mode video, performance metrics on fetal imaging lagged behind adult; thus the performance differential between real vs synthetic B-mode data was not statistically significant.

Additionally, the pipeline requires preprocessing with a U-Net to generate CFD masks. In the future, diffusion-based models may offer improved performance, as they have recently outperformed GANs in image processing, although often at the expense of greater training data requirements. Future work can also add additional modules and loss terms in order to learn and attend to CFD regions within the end-to-end pipeline, eliminating the preprocessing step.

With additional training on more, diverse ultrasound data, and with more testing across additional datasets and structures in the future, such a model can serve as a foundation for video translation for ultrasound and could be expanded to further sub-modalities and super-resolution tasks.

## Methods

*Data.* De-identified ultrasound imaging data from UCSF was used, with waived consent in compliance with the Institutional Review Board. An ultrasound exam includes numerous images displaying a variety of sub-modalities; for this study, frames containing a simultaneous collection of both B-mode and CFD sub-modalities—so-called 'dual' B-mode/CFD acquisitions— were used for training. We created 10-frame long video snippets, with no overlap. This resulted in 2.25 ± 0.7 (1–15) videos per clip for adult (covering a full cardiac cycle), and 13 ± 10 (1–90) videos per clip for fetal studies (using all available frames).

The adult cardiac ultrasound dataset comprised 21,949 such videos from 982 exams. The fetal ultrasound dataset comprised 41427 videos from 465 exams; this included both fetal screening ultrasound studies as well as fetal cardiac ultrasound. Both datasets included studies from a range of patients and pathologies, and videos from a range of different views and angles (**Table S1**). Adult views included: apical two chamber (A2C), apical five chamber (A5C), apical three chamber/long axis (A3C), four chamber (4CH), abdominal aorta or subcostal inferior vena cava (AortaIVC), aortic arch or superior vena cava (AortaSVC), parasternal long axis (PLAX), right ventricular inflow (RVI), right ventricular outflow (RVO), short axis at mid or mitral level (SAX), basal short axis (SAXB), and subcostal four-chamber (SUB4C), as previously described[4]. Fetal views included: three-vessel trachea (3VT), three-vessel view (3VV), left-ventricular outflow tract (LVOT), four chamber (4CH), abdomen (ABDO), and non-target (NT), as previously described[5].

This resulted in an overall dataset of 63376 videos from 1447 exams; about 87% of videos were used for training/validation (55006 videos, 1291 exams), while the remaining 13% (8370 videos, 156 exams) were used for testing. Data in the training/validation and test sets did not overlap by video, clip, or study.

*Preprocessing.* Dual ultrasound RGB frames were split into their constituent B-mode and CFD regions; these were padded to square aspect ratio and resized to 256x256 pixels using Python's OpenCV library.

*Estimating CFD masks.* To focus the model on CFD regions, we identified and quantified the CFD coverage within each video, using this information to (i) generate binary masks (**Figs. S1D, S2B**) for use in model training (**Fig. 2**). A helper U-Net model was trained to segment CFD masks from input CFD frames (**Fig. S1**). The proportion of foreground pixels in each CFD mask was calculated across the training set and data was accordingly divided into quartiles (Q), serving as a quantitative measure of CFD coverage. CFD coverage was first estimated per-frame and then averaged per video.

*Model architectures and training.* We adopt a coarse-to-fine approach as follows, where the coarse network provides inpainting and reconstruction of anatomic structures and the fine network refines video style. To address the limitations of standard convolutions, which apply the same filters to both B-mode (valid) and CFD (invalid) pixels leading to blurriness and unnatural edges, we leveraged gated convolutions[25,27,30] in the coarse network (**Fig. 2A**). Gated convolutions learn a gating mechanism to adaptively select spatially relevant features[24]; our model was trained using L1 loss for 20 epochs prior to training the refinement GAN. While effective for semantic consistency, gated convolutions suppress high-frequency content, resulting in a blurry output (see coarse example, **Fig. S2**)[25,30].

To enhance subtle edges, we post-processed the coarse output before passing it to the refinement network. This offline step includes adding speckle noise (0±0.025), stretching the intensity histogram to enhance contrast (Pillow.ImageEnhance; factor 1.2), and applying edge-enhancing convolution filters (Pillow.ImageFilter.DETAIL, 2 iterations) for sharpening. Additional edge enhancement was applied at inference time to improve perceptual sharpness. In particular, edge enhancement (Pillow.ImageFilter.EDGE_ENHANCE) and contrast enhancement (Pillow.ImageEnhance; factor 1.5) algorithms were applied within a blurred version of the CFD mask (cv2.GaussianBlur, kernel size 25; sigma 50) prior to the pre-processing steps above (see post-processed coarse example, **Fig. S2**).

The refinement network architecture is shown in **Fig. 2B**; each block includes two convolutional layers, followed by instance normalization and a ReLU activation.

The number of filters was set to 32-256, and 16-256 range (both doubling at each layer) for coarse and refinement networks respectively; kernel size was 3. MaxPooling was used for downsampling; with size (2,2,2) in the first block and (1,2,2) otherwise, upsampling stride was therefore (2,2,2) in the last convolutional block, and (1,2,2) otherwise. The total number of parameters was 7738146 for coarse and 10905761 for refinement network.

We adopted 3D spatially-strided convolution in the discriminator [26] (stride size (1,2,2)) with kernel size (3, 5, 5) and LeakyReLU activation function; spectral normalization was implemented. Number of the filters was set to 8-64, doubling at each layer. The discriminator's architecture is shown in **Fig. 2B**.

The refinement network was trained with the overall loss function defined as:

$$L_{total} = \lambda_{l1} L_1 + \lambda_{l1mask} L_{1mask} + \lambda_{perceptual} L_{perceptual} + \lambda_{adversarial} L_{adversarial}$$

Where $\lambda_{l1} = 5$; $\lambda_{l1mask} = 20$; $\lambda_{perceptual} = 0.1$; $\lambda_{l1adversarial} = 1$. L1 mask refers to the previously defined CFD mask, used to assign higher penalty to CFD regions[27,30]. Perceptual loss, commonly used in image generation and inpainting to reduce blurriness[24,27,29,30], was implemented using the discriminator's weights. Validation SSIM was monitored and training was stopped after its value stabilized (reached 0.93) and generated videos demonstrated visually satisfying quality (50 epochs).

For each model, training was conducted on an NVIDIA L40S GPU with a batch size of 16. We employed the Adam optimizer with learning rates of 0.0005, 0.0001, and 0.0002 for the coarse, refinement, and discriminator models respectively. Betas were set at 0.9 and 0.999, and half precision was deployed.

*Data augmentation.* We used estimated the quartiles of CFD coverage and assigned weights (1, 2, 3, or 4 for Q1, Q2, Q3, and Q4, respectively) to prioritize videos with higher CFD coverage during training (PyTorch.WeightedRandomSampler).

Additionally, the following data augmentations were applied during training: random rotation within ±10° (p=0.35), horizontal and vertical flips (p=0.35 each), random shifts up to 10% (p=0.35), and random scaling up to 8% (p=0.35). Gaussian blur (p=0.35) and additive Gaussian noise (0±0.03, p=0.35) were applied for CFD input only.

*DL model evaluation of synthetic videos.* Previously developed view classification, segmentation models and biometrics for the adult and fetal data, respectively[1,4,6–8], were used to infer on ground-truth B-mode frames as well as synthetic frames. The final view classification prediction for each 10-frame video was determined by the most frequently predicted view. The view classifier's performance was evaluated by the normalized confusion matrix, overall F1 score, and per-class F1 scores. Segmentation performance was evaluated on frames only, using the Dice scores.

*Blinded clinical expert evaluation of synthetic videos.* We selected 200 video examples (ground-truth B-mode video and corresponding synthetic video) each from adult and fetal datasets for evaluation by clinical experts board-certified or board-eligible in adult or fetal ultrasound, respectively (10.9±6.7 years experience, range 1-20). 100 examples were used for each of the following tests.

In the first test, blinded clinicians were asked to classify videos by view. Unknown to the clinicians, approximately half the videos were real and half were synthetic. After the fact, clinician performance on real vs synthetic videos was analyzed. All clinicians took the first test.

Then, half the clinicians took the second test and half took the third test. In the second test, clinicians were shown one video and asked to choose if it was real or synthetic. In the third test, clinicians were presented with a pair of videos—the real B-mode video and its synthetic counterpart—and asked to choose the real video. It was only in the second and third tests that the clinicians may have been prompted that the study was about video synthesis (due to the nature of the task).

Finally, the same videos were used in the second test, but this time only synthetic were given to non-blinded subjects (clinical and non-clinical) to evaluate for realism. For each sample, a fool rate coefficient was computed as the proportion of subjects who perceived it as real. We then calculated the correlation between the fool rate and the SSIM scores comparing B-mode and synthetic images.

Notably, GANs are not well-suited for generating text or static fiducials in videos[38,39]. Therefore, we replaced these elements in the synthetic outputs with the corresponding fiducials from the real B-mode data. We used the Segment Anything Model (SAM-Huge[40]) to detect and extract the fiducials.

*Statistical testing.* Statistical tests were performed using the two-sample, non-parametric Mann-Whitney U (MWU) test except where a different test is mentioned.

## References

1. Ferreira, D., Lau, C., Salaymang, Z. & Arnaout, R. Self-supervised learning for label-free segmentation in cardiac ultrasound. *Nat Commun* **16**, 4070 (2025).
2. Dabiri, Y. *et al.* Mitral Valve Atlas for Artificial Intelligence Predictions of MitraClip Intervention Outcomes. *Front Cardiovasc Med* **8**, 759675 (2021).
3. Datar, Y. *et al.* Myocardial Texture Analysis of Echocardiograms in Cardiac Transthyretin Amyloidosis. *J Am Soc Echocardiogr* **37**, 570–573 (2024).
4. Madani, A., Arnaout, R., Mofrad, M. & Arnaout, R. Fast and accurate view classification of echocardiograms using deep learning. *npj Digital Med* **1**, 1–8 (2018).
5. Holste, G. *et al.* Complete AI-Enabled Echocardiography Interpretation With Multitask Deep Learning. *JAMA* **334**, 306–318 (2025).
6. Arnaout, R. *et al.* An ensemble of neural networks provides expert-level prenatal detection of complex congenital heart disease. *Nat Med* **27**, 882–891 (2021).
7. Athalye, C. *et al.* Deep-learning model for prenatal congenital heart disease screening generalizes to community setting and outperforms clinical detection. *Ultrasound Obstet Gynecol* **63**, 44–52 (2024).
8. Kornblith, A. E. *et al.* Development and Validation of a Deep Learning Strategy for Automated View Classification of Pediatric Focused Assessment With Sonography for Trauma. *J Ultrasound Med* **41**, 1915–1924 (2022).
9. Reddy, A., Rizvi, S., Moon-Grady, A. J. & Arnaout, R. Improving Prenatal Detection of Congenital Heart Disease With a Scalable Composite Analysis of 6 Fetal Cardiac Ultrasound Biometrics. *J Am Soc Echocardiogr* **37**, 1186–1188 (2024).
10. Arnaout, R. Can Machine Learning Help Simplify the Measurement of Diastolic Function in Echocardiography? *JACC Cardiovasc Imaging* **14**, 2105–2106 (2021).

11. Chinn, E., Arora, R., Arnaout, R. & Arnaout, R. ENRICHing medical imaging training sets enables more efficient machine learning. *J Am Med Inform Assoc* ocad055 (2023) doi:10.1093/jamia/ocad055.

12. Ferreira, D. & Arnaout, R. Are foundation models efficient for medical image segmentation? *Journal of the American Society of Echocardiography* **38**, 514–516 (2025).

13. Arnaout, R. Adapting vision-language AI models to cardiology tasks. *Nat Med* **30**, 1245–1246 (2024).

14. Sachdeva, R. *et al.* Novel Techniques in Imaging Congenital Heart Disease. *JACC* **83**, 63–81 (2024).

15. Miao, B. Y. *et al.* The MI-CLAIM-GEN checklist for generative artificial intelligence in health. *Nat Med* **31**, 1394–1398 (2025).

16. Dey, D. *et al.* Proceedings of the NHLBI Workshop on Artificial Intelligence in Cardiovascular Imaging: Translation to Patient Care. *JACC: Cardiovascular Imaging* **16**, 1209–1223 (2023).

17. Arnaout, R. ChatGPT Helped Me Write This Talk Title, but Can It Read an Echocardiogram? *Journal of the American Society of Echocardiography* **0**, (2023).

18. Couch, J., Arnaout, R. & Arnaout, R. Beyond Size and Class Balance: Alpha as a New Dataset Quality Metric for Deep Learning. *ArXiv* arXiv:2407.15724v2 (2024).

19. Nguyen, P. *et al.* $\textit{greylock}$: A Python Package for Measuring The Composition of Complex Datasets. Preprint at https://doi.org/10.48550/arXiv.2401.00102 (2023).

20. Athalye, C. & Arnaout, R. Domain-guided data augmentation for deep learning on medical imaging. *PLoS One* **18**, e0282532 (2023).

21. AIUM Curriculum for Fundamentals of Ultrasound Physics and Instrumentation. *Journal of Ultrasound in Medicine* **38**, 1933–1935 (2019).

22. Islam, K. T. *et al.* Improving portable low-field MRI image quality through image-to-image translation using paired low- and high-field images. *Sci Rep* **13**, 21183 (2023).

23. Yang, Q. *et al.* MRI Cross-Modality Image-to-Image Translation. *Sci Rep* **10**, 3753 (2020).

24. Armanious, K. *et al.* MedGAN: Medical image translation using GANs. *Computerized Medical Imaging and Graphics* **79**, 101684 (2020).

25. Yu, J. *et al.* Free-Form Image Inpainting With Gated Convolution. in *2019 IEEE/CVF International Conference on Computer Vision (ICCV)* 4470–4479 (2019). doi:10.1109/ICCV.2019.00457.

26. Iizuka, S., Simo-Serra, E. & Ishikawa, H. Globally and locally consistent image completion. *ACM Trans. Graph.* **36**, 1–14 (2017).

27. Chang, Y.-L., Liu, Z. Y. & Lee, K.-Y. Free-Form Video Inpainting With 3D Gated Convolution and Temporal PatchGAN. in *2019 IEEE/CVF International Conference on Computer Vision (ICCV)* 9065–9074 (IEEE, Seoul, Korea (South), 2019). doi:10.1109/ICCV.2019.00916.

28. Image Inpainting for Irregular Holes Using Partial Convolutions. in *Lecture Notes in Computer Science* 89–105 (Springer International Publishing, Cham, 2018). doi:10.1007/978-3-030-01252-6_6.

29. Armanious, K., Mecky, Y., Gatidis, S. & Yang, B. Adversarial Inpainting of Medical Image Modalities. in *ICASSP 2019 - 2019 IEEE International Conference on Acoustics, Speech and Signal Processing (ICASSP)* 3267–3271 (2019). doi:10.1109/ICASSP.2019.8682677.

30. Li, B. *et al.* AIVUS: Guidewire Artifacts Inpainting for Intravascular Ultrasound Imaging With United Spatiotemporal Aggregation Learning. *IEEE Transactions on Computational Imaging* **8**, 679–692 (2022).

31. Shmelkov, K., Schmid, C. & Alahari, K. How Good Is My GAN? in *Computer Vision – ECCV 2018* (eds Ferrari, V., Hebert, M., Sminchisescu, C. & Weiss, Y.) 218–234 (Springer International Publishing, Cham, 2018). doi:10.1007/978-3-030-01216-8_14.

32. Isola, P., Zhu, J.-Y., Zhou, T. & Efros, A. A. Image-to-Image Translation with Conditional Adversarial Networks. in *2017 IEEE Conference on Computer Vision and Pattern Recognition (CVPR)* 5967–5976 (2017). doi:10.1109/CVPR.2017.632.

33. Johnson, J., Alahi, A. & Fei-Fei, L. Perceptual Losses for Real-Time Style Transfer and Super-Resolution. Preprint at https://doi.org/10.48550/arXiv.1603.08155 (2016).

34. Matta, S. *et al.* A systematic review of generalization research in medical image classification. *Computers in Biology and Medicine* **183**, 109256 (2024).

35. Breger, A. *et al.* A Study of Why We Need to Reassess Full Reference Image Quality Assessment with Medical Images. *J Digit Imaging. Inform. med.* https://doi.org/10.1007/s10278-025-01462-1 (2025) doi:10.1007/s10278-025-01462-1.

36. Pambrun, J.-F. & Noumeir, R. Limitations of the SSIM quality metric in the context of diagnostic imaging. in *2015 IEEE International Conference on Image Processing (ICIP)* 2960–2963 (2015). doi:10.1109/ICIP.2015.7351345.

37. Maani, F. *et al.* FetalCLIP: A Visual-Language Foundation Model for Fetal Ultrasound Image Analysis. Preprint at https://doi.org/10.48550/arXiv.2502.14807 (2025).

38. Rodriguez, J. A., Vazquez, D., Laradji, I., Pedersoli, M. & Rodriguez, P. OCR-VQGAN: Taming Text-within-Image Generation. in *2023 IEEE/CVF Winter Conference on Applications of Computer Vision (WACV)* 3678–3687 (IEEE, Waikoloa, HI, USA, 2023). doi:10.1109/WACV56688.2023.00368.

39. Fang, S., Xie, H., Chen, J., Tan, J. & Zhang, Y. Learning to Draw Text in Natural Images with Conditional Adversarial Networks. in *Proceedings of the Twenty-Eighth International Joint Conference on Artificial Intelligence* 715–722 (International Joint Conferences on Artificial Intelligence Organization, Macao, China, 2019). doi:10.24963/ijcai.2019/101.

40. Kirillov, A. *et al.* Segment Anything. Preprint at https://doi.org/10.48550/arXiv.2304.02643 (2023).

## Author contributions

R.A. conceived of the study. R.A., N.T., and S.J. designed experiments with input from T.Y.L. N.T. performed experiments and analyzed data with help from R.B., L.G., S.J., T.Y.L., and C.L. N.T. and R.A. wrote the manuscript with input from all authors.

**Acknowledgements**

We thank Kami Gill, Claudia Guitierrez, Roshana Goodar, Shari Kennedy, Megan McLaughlin, Jane Glover, Vanessa Flores, Evette Iweke, Carol Leung, Caitlyn Brenner and other clinicians and sonographers who wish to remain anonymous, who served as expert evaluators of imaging. We thank Lennart Elbe for insightful suggestions on the manuscript.

**Data availability statement:** Code will be made available upon publication.

**Competing Interests:** None.

## Tables

**Table 1. Performance of clinical experts on three tasks and three datasets.**

| Task | Dataset | Data origin | Accuracy, mean±std (range) | P value |
|---|---|---|---|---|
| **Single video: "Is it real?"** | Adult | Real or synthetic | 54±6 (46-61) | — |
| | Fetal | Real or synthetic | 56.7±2.5 (53.1-60.2) | — |
| | Other anatomies | Real or synthetic | 51.9±9.9 (41.4-65.3) | — |
| **Pair of videos: "Choose real."** | Adut | Real and synthetic | 46.5±4.5 (41.9-52.6) | — |
| | Fetal | Real and synthetic | 55.2±4.9 (48.9-61.2) | — |
| | Other anatomies | Real and synthetic | 78.9±5 (71.7-83.9) | — |
| **Task** | **Dataset** | **Data origin** | **F1, mean±std (range)** | **P value** |
| **Single video: "Classify by view."** | Adult | Real* | 0.73±0.10 (0.60-0.88) | 0.26 |
| | | Synthetic* | 0.81±0.11 (0.66-0.91) | |
| | Fetal | Real* | 0.54±0.12 (0.41-0.74) | 0.27 |
| | | Synthetic* | 0.46±0.07 (0.35-0.54) | |
| | Other anatomies | Real* | 0.71±0.1 (0.51-0.82) | 0.06 |
| | | Synthetic* | 0.82±0.05 (0.74-0.87) | |

* Clinicians were shown a mixture of real and synthetic images in a blinded fashion, analysis was performed post-hoc.

## Figures

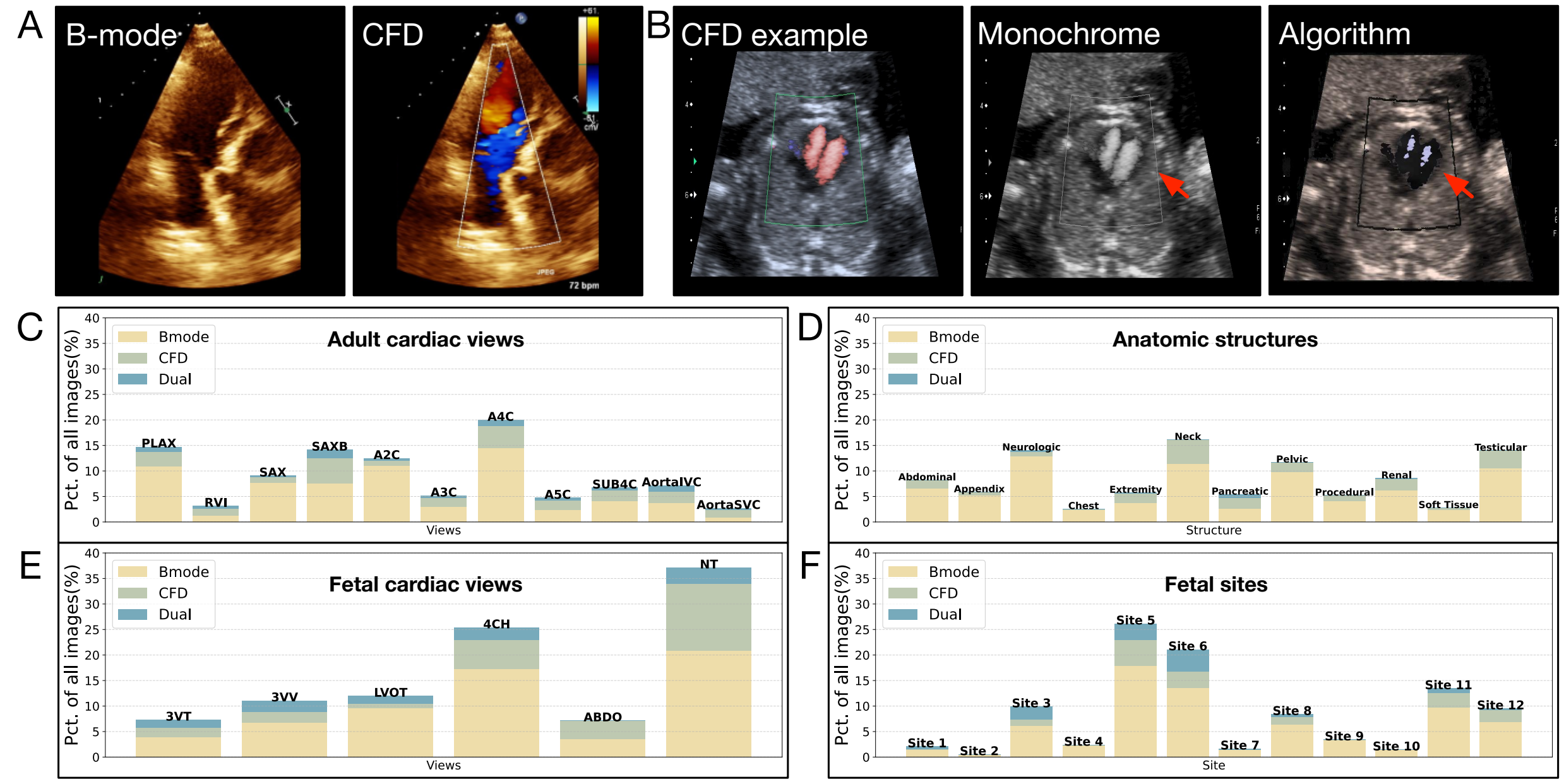

**Figure 1. B-mode and color flow doppler (CFD) offer different information, are unbalanced in datasets, and are not easy to translate. (A)** B-mode and CFD are often used interchangeably, but they in fact carry different information. Of note, here, both B-mode and CFD videos use an orange chromamap. **(B)** Translation between CFD and B-mode is not easily accomplished using simple algorithms. Monochrome rendering makes the CFD signal grey (red arrow) instead of matching the blood pool. Algorithmically detecting colored pixels and replacing them with grayscale values has several problems, including retaining the CFD region of interest box, imperfect thresholds for colored pixels and replacement greyscale values that do not match the color or noise profile of the ultrasound image at hand (red arrow). Furthermore, anatomic structures underneath the original color signal are not reconstructed. The prevalence of B-mode, CFD, and dual acquisitions in ultrasound imaging is imbalanced and varies across typical clinical views **(C, E)**; different anatomic structures **(D)**; and even within the same structure dataset across different clinical sites, designated here as Sites 1-12 **(F)**. PLAX – parasternal long axis, RVI – right ventricular inflow, SAX – short axis at mid or mitral level, SAXB – basal short axis, SAX – short axis at mid or mitral level, SAXB – basal short axis, A2C – apical two chamber, A3C – apical three chamber/long axis, 4CH – four chamber, A5C – apical five chamber, SUB4C – subcostal four-chamber, AortaIVC – abdominal aorta or subcostal inferior vena cava, AortaSVC – aortic arch or superior vena cava, 3VT – three-vessel trachea, 3VV – three-vessel view, LVOT – left-ventricular outflow tract, ABDO – abdomen, NT – non-target.

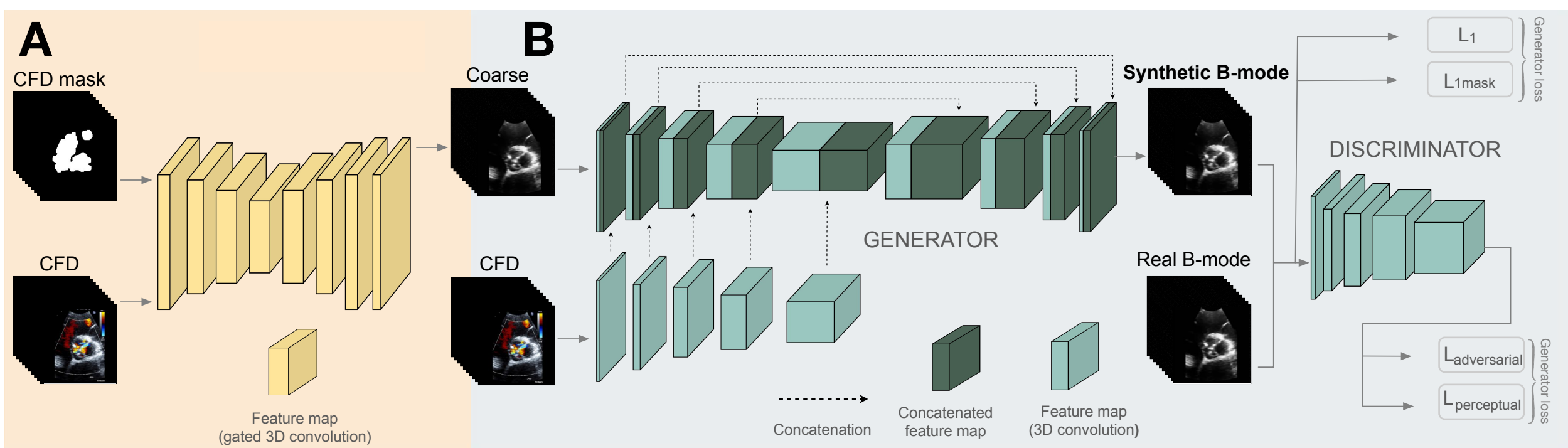

**Figure 2. Coarse-to-fine GAN provides video-to-video translation while preserving anatomical content and realistic style. (A)** The coarse network inputs the CFD video and a mask of the CFD region and uses gated convolutions to adaptively leverage spatially relevant features, producing coherent outputs. **(B)** The refinement GAN combines L1 loss with adversarial and perceptual losses derived from the discriminator (see Methods for more details).

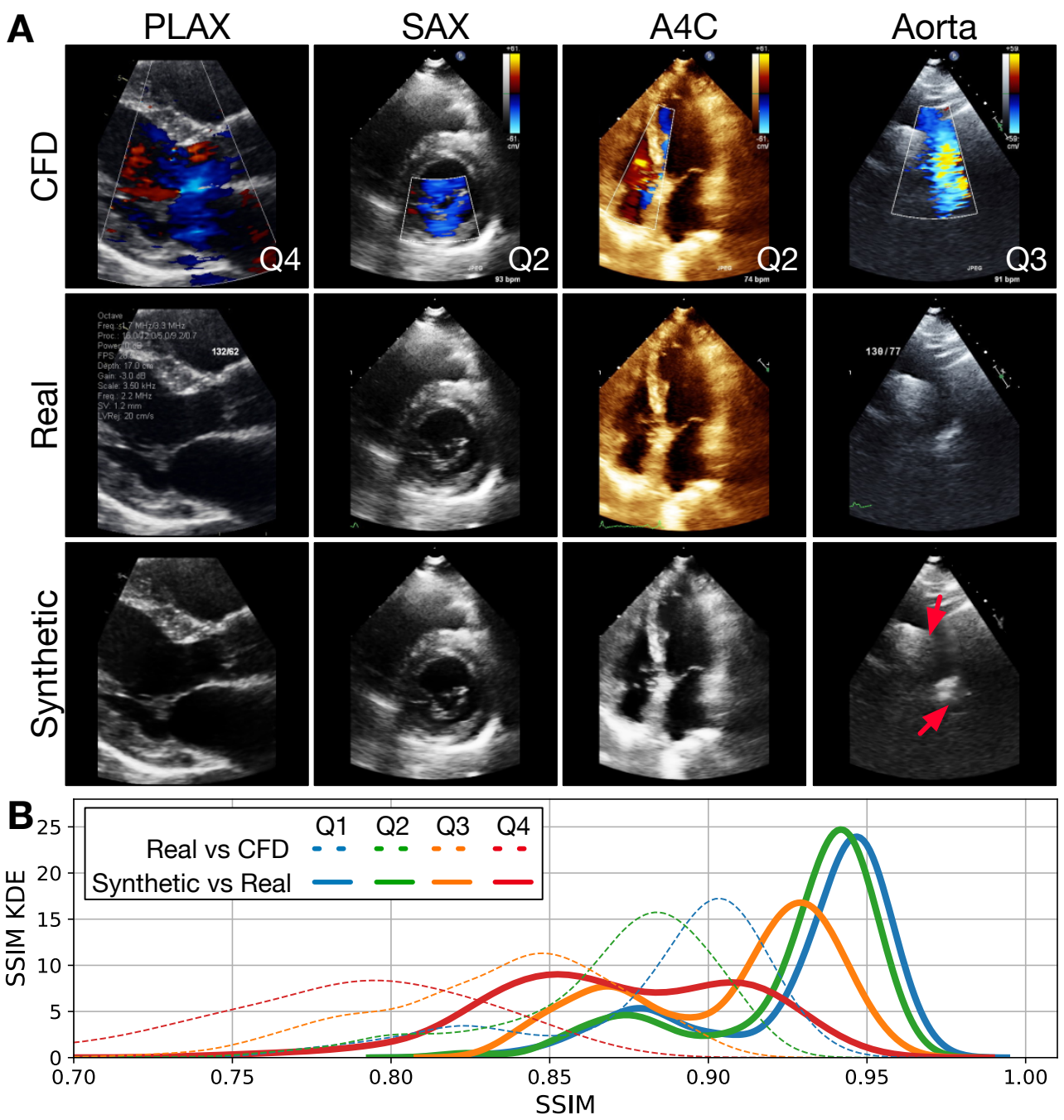

**Figure 3. GAN pipeline can be trained to generate realistic synthetic B-mode videos with high structural similarity for adult ultrasound. (A)** Example adult ultrasound views showing CFD, corresponding real (ground-truth) B-mode, and synthetic B-mode video frames. Note the synthetic frames are realistic despite the orange chromamap used in the apical 4-chamber (A4C) example, and anatomic structures underneath the color signal can be faithfully reconstructed (e.g. red arrows) across a range of CFD coverages (quartiles of coverage Q1, Q2, Q3, and Q4 appear in lower right-hand corner of each CFD example). **(B)** SSIM kernel density estimates (KDE) comparing two SSIM distributions: B-mode (ground truth) to CFD input data (dashed line) and B-mode to synthetic data (solid line). Q1 represents videos with the lowest CFD coverage (see Methods), and Q4 the highest. SSIM between synthetic and ground-truth B-mode videos is high overall, although slightly lower in the quartile of videos with the most CFD signal. Across the board, similarity of synthetic to ground-truth videos is higher than that of ground-truth B-mode compred to the original CFD image, indicating successful video translation (see text and **Table S3** for statistical tests). SAX-short axis at mitral valve level, PLAX-parasternal long axis, 4CH-four chamber view, Aorta-aortic arch view.

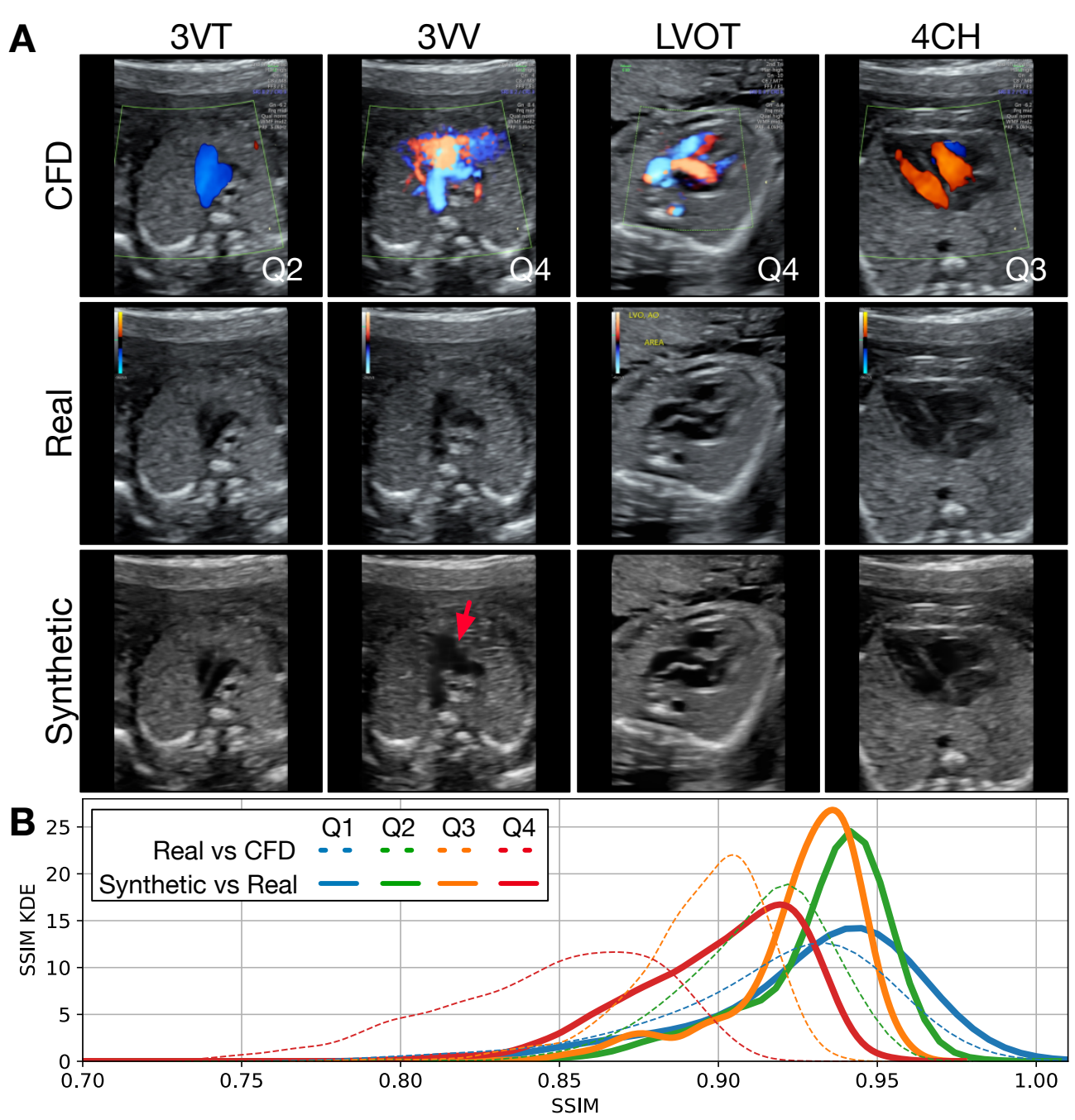

**Figure 4. GAN pipeline also generates realistic B-mode videos for fetal ultrasound. (A)** Examples of CFD, corresponding real (ground-truth) B-mode, and synthetic B-mode video frames for fetal ultrasound. Quartiles of CFD coverage are noted in lower right-hand corner of each CFD example). Note that due to the much smaller heart in fetal ultrasound, the CFD signal tends to obfuscate the anatomic structures of interest much more than in adult, making video translation a more challenging task. Consequently, some fine structures, such as the septation between the pulmonary artery and the aorta in the 3VV view (red arrow), are less clear in the synthetic video compared to ground truth. **(B)** SSIM kernel density estimates (KDE) comparing B-mode (ground truth) to CFD input data (dashed line) and B-mode to synthetic data (solid line). Again, SSIM between synthetic and ground-truth B-mode videos is high, and similarity of synthetic to ground-truth videos is higher than that of ground-truth B-mode compared to the original CFD image (see text and **Table S3** for statistical tests). 3VT-3-vessel trachea, LVOT-left-ventricular outflow tract, 3VV-3-vessel view.

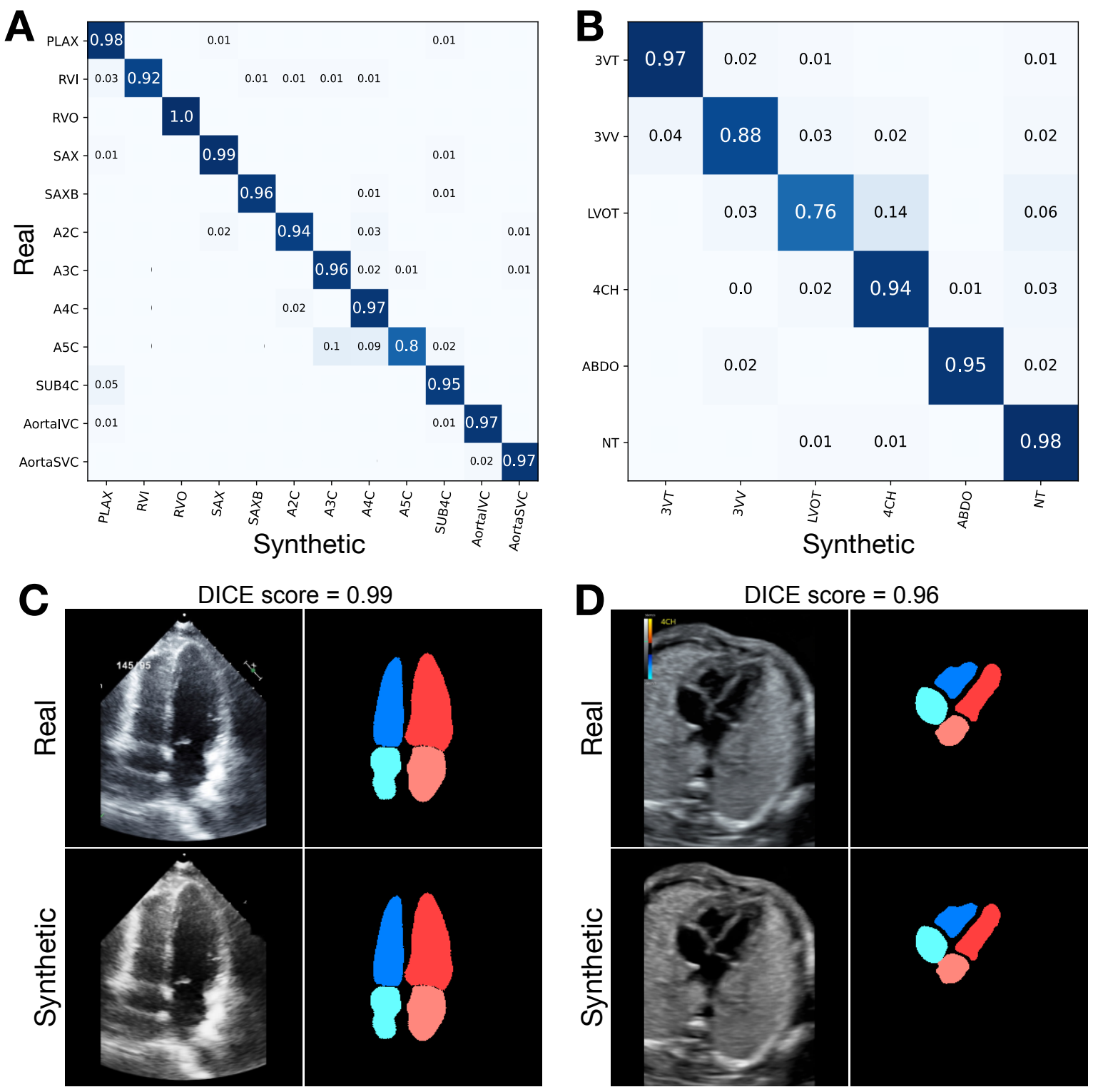

**Figure 5. Functional testing of synthetic videos in benchmark deep learning tasks demonstrate good performance compared to real B-mode videos. (A-B)** Confusion matrices comparing adult (A) and fetal (B) view classification on real vs synthetic B-mode videos. **(C-D)** Examples of semantic segmentation on real vs synthetic B-mode frames from adult (C) and fetal (D) datasets. PLAX – parasternal long axis, RVI – right ventricular inflow, RVO – right ventricular outflow, SAX – short axis at mid or mitral level, SAXB – basal short axis, SAX – short axis at mid or mitral level, SAXB – basal short axis, A2C – apical two chamber, A3C – apical three chamber/long axis, 4CH – four chamber, A5C – apical five chamber, SUB4C – subcostal four-chamber, AortaIVC – abdominal aorta or subcostal inferior vena cava, AortaSVC – aortic arch or superior vena cava, 3VT – three-vessel trachea, 3VV – three-vessel view, LVOT – left-ventricular outflow tract, ABDO – abdomen, NT – non-target.

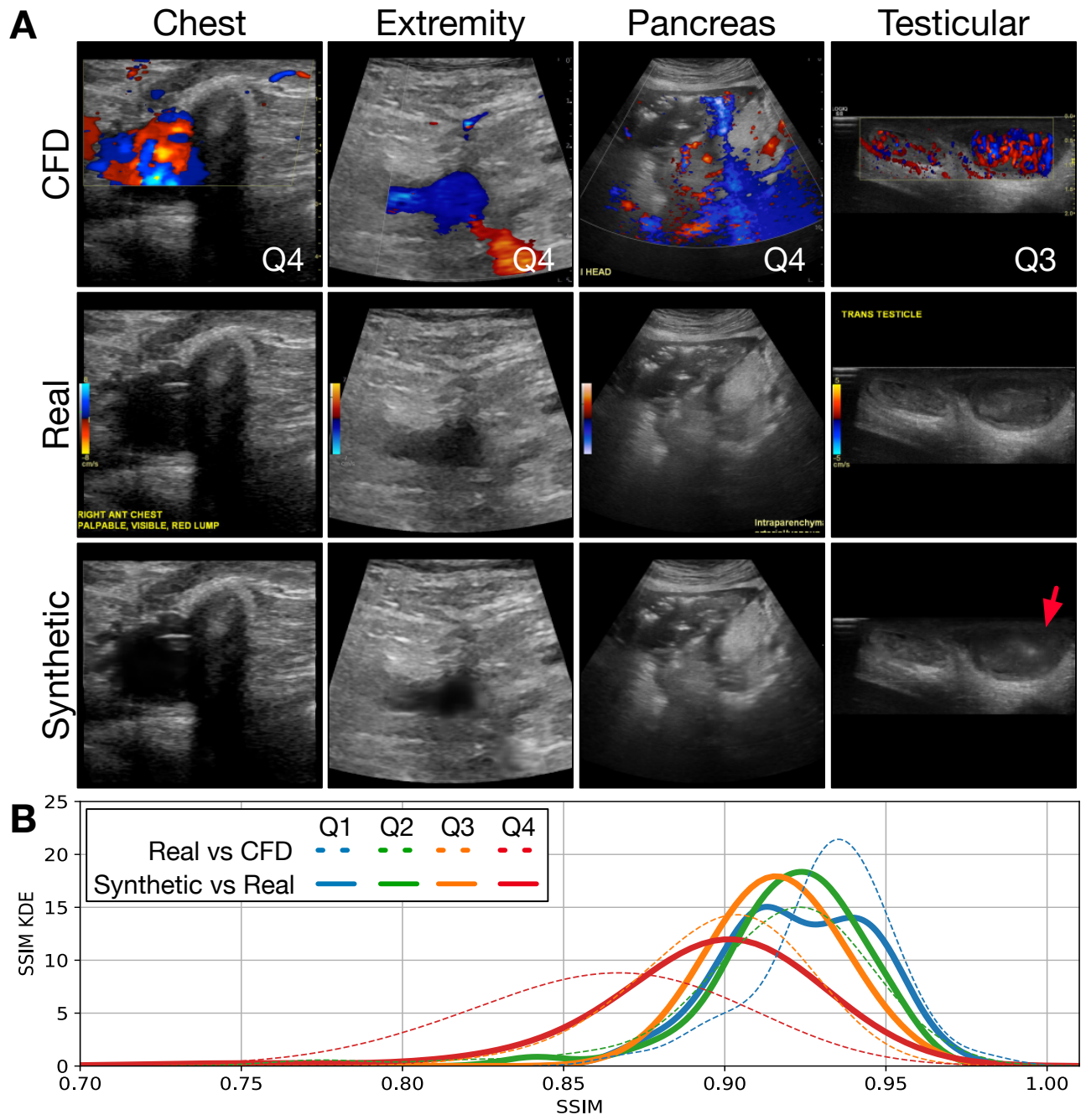

**Figure 6. Zero-shot performance on several anatomic structures unseen during model training produces similarly realistic synthetic videos, suggesting foundational capabilities for video translation. (A)** Examples of CFD frames, corresponding real (ground-truth) B-mode frames, and synthetic B-mode frames across various anatomic structures not represented in model training. The underlying anatomy is mostly reconstructed, but some artifacts are observed (red arrow). **(B)** SSIM kernel density estimates (KDE) comparing B-mode (ground truth) to CFD input data (dashed line) and B-mode to synthetic data (solid line). CFD videos are divided into quartiles (Q) of CFD coverage.

## Supplemental Methods

*S1 Helper U-Net.* To generate binary masks of the CFD-containing regions for the coarse network, we trained a helper U-Net to detect the CFD region from each frame. To train this model, we leveraged the subset of dual B-mode-CFD frames (Fig. S1 A).

We generated training labels as follows. First, we subtracted the B-mode portion of the dual frame from the CFD portion (Fig. S1A) (Pillow version 11.1.0, ImageChops). We then segmented the Doppler color sector (white arrow, Fig. S1A) using the Segment Anything Model (SAM-Huge[35]) and used this (Fig. S1C) to mask out annotations (red arrows, Fig. S1A)), fiducial markings and color bar where applicable. This resulted in usable labels for the CFD region only (Fig. S1D).

We then trained a custom U-Net with CFD (Fig. S1A) as input and the prepared labels (n=1120) as outputs. Dilation-erosion operations (OpenCV version 4.11.0.86, dilate, erode) were used to remove any residual color doppler box. The final CFD masks are shown on Fig. S2.

## Supplemental Tables

### Table S1. Training, validation, and test data.

| | Train | | Validation | | Test | |
|---|---|---|---|---|---|---|
| **View** | **Videos** | **Studies** | **Videos** | **Studies** | **Videos** | **Studies** |
| **A2C** | 765 (4.8%) | 267 (7.1%) | 138 (4.7%) | 56 (8.0%) | 137 (4.8%) | 49 (7.2%) |
| **A5C** | 1093 (6.8%) | 323 (8.5%) | 196 (6.7%) | 58 (8.2%) | 243 (8.5%) | 64 (9.4%) |
| **A3C** | 733 (4.6%) | 237 (6.3%) | 145 (4.9%) | 44 (6.2%) | 149 (5.2%) | 47 (6.9%) |
| **A4C** | 2878 (17.9%) | 431 (11.4%) | 553 (18.8%) | 84 (11.9%) | 557 (19.4%) | 82 (12.1%) |
| **AortaIVC** | 1796 (11.2%) | 486 (12.9%) | 335 (11.4%) | 92 (13.1%) | 282 (9.8%) | 78 (11.5%) |
| **AortaSVC** | 1316 (8.2%) | 352 (9.3%) | 275 (9.3%) | 72 (10.2%) | 241 (8.4%) | 58 (8.5%) |
| **PLAX** | 1120 (7.0%) | 292 (7.7%) | 189 (6.4%) | 50 (7.1%) | 241 (8.4%) | 55 (8.1%) |
| **RVI** | 671 (4.2%) | 245 (6.5%) | 105 (3.6%) | 43 (6.1%) | 89 (3.1%) | 39 (5.7%) |
| **RVO** | 41 (0.3%) | 21 (0.6%) | 8 (0.3%) | 5 (0.7%) | 9 (0.3%) | 4 (0.6%) |
| **SAX** | 786 (4.9%) | 304 (8.0%) | 134 (4.5%) | 54 (7.7%) | 147 (5.1%) | 62 (9.1%) |
| **SAXB** | 3081 (19.1%) | 447 (11.8%) | 536 (18.2%) | 76 (10.8%) | 523 (18.3%) | 80 (11.8%) |
| **SUB4C** | 1825 (11.3%) | 375 (9.9%) | 333 (11.3%) | 70 (9.9%) | 246 (8.6%) | 61 (9.0%) |
| **Unique adult** | 16105 | 728 | 2947 | 134 | 2864 | 120 |
| **3VT** | 1485 (5.2%) | 164 (13.5%) | 396 (5.3%) | 36 (13.3%) | 314 (5.7%) | 24 (14.8%) |
| **3VV** | 2207 (7.8%) | 198 (16.3%) | 429 (5.7%) | 43 (15.9%) | 488 (8.9%) | 24 (14.8%) |
| **LVOT** | 4466 (15.7%) | 233 (19.2%) | 1325 (17.7%) | 53 (19.6%) | 929 (16.9%) | 29 (17.9%) |
| **4CH** | 3091 (10.9%) | 216 (17.8%) | 913 (12.2%) | 49 (18.1%) | 1139 (20.7%) | 32 (19.8%) |
| **ABDO** | 348 (1.2%) | 73 (6.0%) | 119 (1.6%) | 22 (8.1%) | 85 (1.5%) | 18 (11.1%) |
| **NT** | 16830 (59.2%) | 329 (27.1%) | 4314 (57.6%) | 67 (24.8%) | 2549 (46.3%) | 35 (21.6%) |
| **Unique fetal** | 28427 | 358 | 7496 | 71 | 5504 | 36 |
| **Total unique** | 44532 | 1086 | 10443 | 205 | 8368 | 156 |

PLAX – parasternal long axis, RVI – right ventricular inflow, RVO – right ventricular outflow, SAX – short axis at mid or mitral level, SAXB – basal short axis, SAX – short axis at mid or mitral level, SAXB – basal short axis, A2C – apical two chamber, A3C – apical three chamber/long axis, 4CH – four chamber, A5C – apical five chamber, SUB4C – subcostal four-chamber, AortaIVC – abdominal aorta or subcostal inferior vena cava, AortaSVC – aortic arch or superior vena cava, 3VT – three-vessel trachea, 3VV – three-vessel view, LVOT – left-ventricular outflow tract, ABDO – abdomen, NT – non-target.

**Table S2. Additional test data from a range of anatomies not seen during training.**

| Structure | Videos | Studies |
|---|---|---|
| **Extremity** | 11 | 8 |
| **Abdominal** | 15 | 9 |
| **Procedural** | 2 | 1 |
| **Pelvic** | 3 | 2 |
| **Soft Tissue** | 2 | 1 |
| **Neck** | 2 | 1 |
| **Testicular** | 5 | 3 |
| **Chest** | 1 | 1 |
| **Neurologic** | 32 | 17 |
| **Pancreatic** | 147 | 45 |
| **Renal** | 40 | 23 |
| **Total** | 260 | 111 |

**Table S3. SSIM performance by amount of CFD signal per video.**

| | **Synthetic vs real SSIM, mean±std (range)** | **CFD vs real SSIM, mean±std (range)** | **Wasserstein distance** |
|---|---|---|---|
| ADULT Q1 | 0.93±0.03 (0.82, 0.97) | 0.88±0.04 (0.71, 0.94) | 0.05 |
| ADULT Q2 | 0.93±0.03 (0.82, 0.96) | 0.87±0.04 (0.7, 0.93) | 0.06 |
| ADULT Q3 | 0.91±0.03 (0.83, 0.96) | 0.83±0.04 (0.68, 0.91) | 0.07 |
| ADULT Q4 | 0.87±0.05 (0.35, 0.95) | 0.78±0.06 (0.33, 0.87) | 0.09 |
| **OVERALL ADULT** | **0.91 ± 0.04 (0.35, 0.97)** | **0.84 ± 0.06 (0.33, 0.94)** | **0.07** |
| FETAL Q1 | 0.92±0.08 (0.1, 0.98) | 0.91±0.08 (0.11, 0.97) | 0.01 |
| FETAL Q2 | 0.93±0.03 (0.1, 0.98) | 0.91±0.03 (0.11, 0.95) | 0.02 |
| FETAL Q3 | 0.93±0.02 (0.83, 0.96) | 0.89±0.02 (0.78, 0.94) | 0.03 |
| FETAL Q4 | 0.9±0.03 (0.69, 0.96) | 0.85±0.04 (0.62, 0.92) | 0.05 |
| **OVERALL FETAL** | **0.92 ± 0.04 (0.10, 0.98)** | **0.88 ± 0.05 (0.11, 0.97)** | **0.03** |
| STRUCTURES Q1 | 0.92±0.02 (0.87, 0.97) | 0.93±0.02 (0.87, 0.98) | 0.01 |
| STRUCTURES Q2 | 0.92±0.02 (0.84, 0.95) | 0.91±0.03 (0.77, 0.96) | 0.01 |
| STRUCTURES Q3 | 0.91±0.04 (0.59, 0.94) | 0.89±0.04 (0.56, 0.94) | 0.02 |
| STRUCTURES Q4 | 0.89±0.06 (0.43, 0.95) | 0.85±0.07 (0.41, 0.92) | 0.04 |
| **OVERALL STRUCTURES** | **0.91 ± 0.05 (0.43, 0.97)** | **0.89 ± 0.06 (0.41, 0.98)** | **0.02** |
| **OVERALL ALL** | **0.91 ± 0.04 (0.10, 0.98)** | **0.87 ± 0.06 (0.11, 0.98)** | **0.04** |

## Supplemental Figures

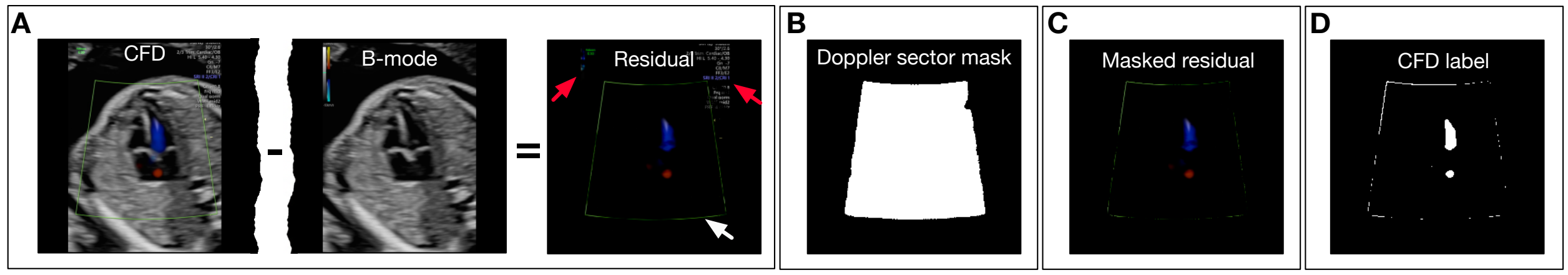

**Figure S1. Generating training labels for helper U-Net. A)** From a dual (CFD/B-mode) acquisition, B-mode is subtracted from CFD and the residual image is kept, which includes the Doppler sector (white arrow) and fiducial markings (red arrows). **B)** The Doppler sector is segmented. **C)** The residual image from (A) is then masked and only the CFD-containing region is kept. **D)** Thresholding is applied to generate usable labels for the CFD-containing region.

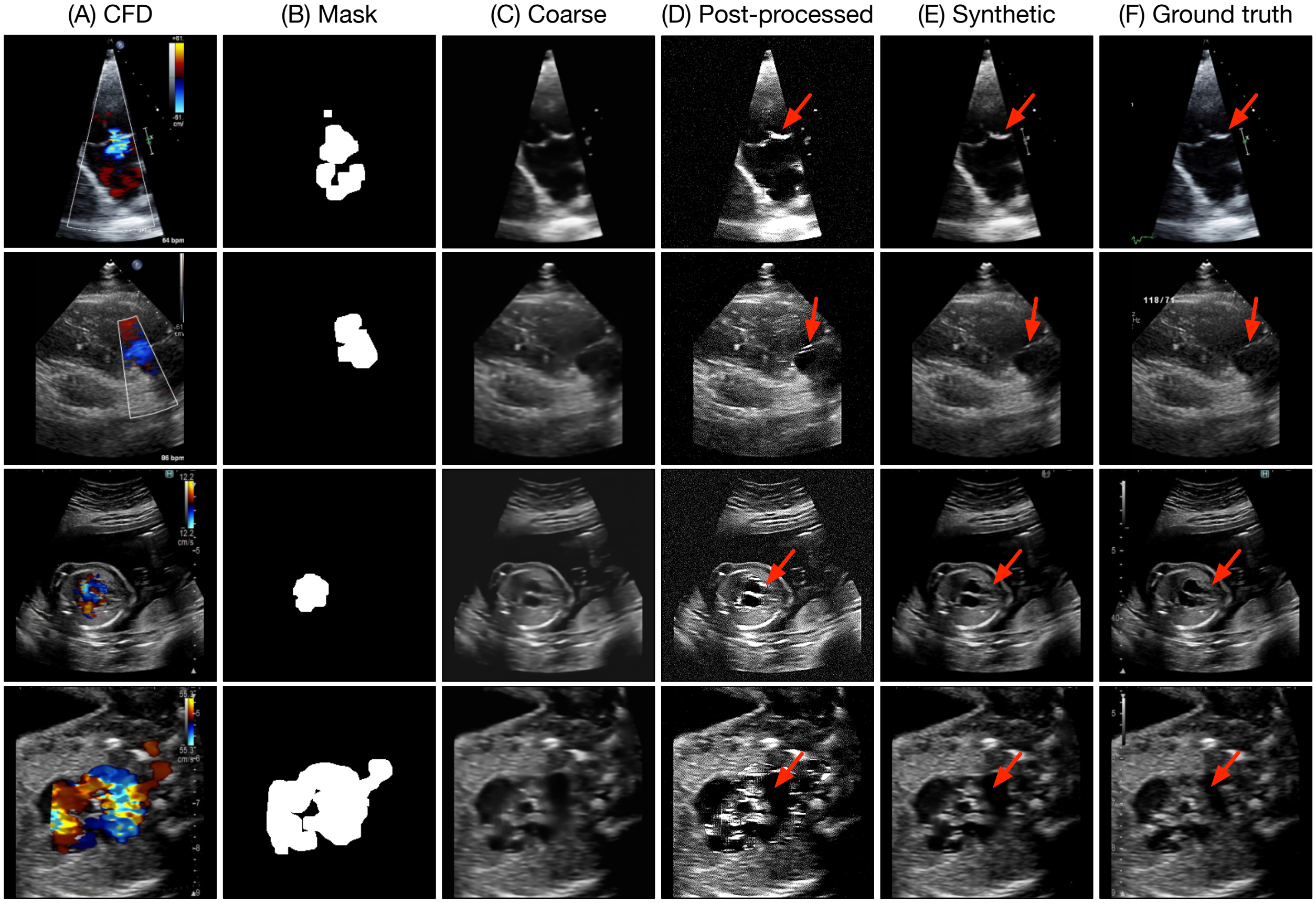

**Figure S2. Examples of CFD input, intermediate, and output frames from steps in the pipeline. (A)** CFD videos are the inputs to the pipeline. **(B)** A CFD mask is generated by helper U-Net and used as input to the coarse network. **(C)** Coarse frame represents the output of the coarse network, which is further post-processed using traditional image processing techniques. **(D)** The post-processed coarse frame may not appear realistic but serves as a conditioning input for the refinement network. Red arrows highlight how post-processing of the coarse output helps emphasize subtle anatomic features, thereby benefiting **(E)** the final synthetic B-mode frame. **(F)** shows example ground-truth frames for comparison.